\documentclass[twoside,11pt]{article}
\usepackage{jmlr2e} 

\usepackage{url}

\usepackage{multirow}

\usepackage{amsmath, amssymb}
\usepackage{bm}

\newcommand{\Rbb}{\mathbb{R}}

\newcommand{\Tbb}{\mathbb{T}}
\newcommand{\Nbb}{\mathbb{N}}

\newcommand{\Ncal}{\mathcal{N}}

\newcommand{\Ocal}{\mathcal{O}}
\newcommand{\Fcal}{\mathcal{F}}

\newcommand{\diff}{\,\text{d}}
\DeclareMathOperator{\diag}{diag}
\DeclareMathOperator{\cond}{cond}

\usepackage{mathtools}
\mathtoolsset{showonlyrefs=true}

\firstpageno{1}

\usepackage{wrapfig}

\renewcommand\vec{\mathbf}
\usepackage{marvosym}

\usepackage{xcolor}

\usepackage{pdfpages}

\usepackage{csvsimple}

\begin{document}

\title{Stable Implementation of Probabilistic ODE Solvers}

\author{\name Nicholas Kr\"amer \email nicholas.kraemer@uni-tuebingen.de \\
\addr University of T\"ubingen\\
Maria-von-Linden-Stra\ss e 6, T\"{u}bingen, Germany
\AND
\name Philipp Hennig \email philipp.hennig@uni-tuebingen.de \\
\addr University of T\"{u}bingen and Max-Planck Institute for Intelligent Systems\\
Maria-von-Linden-Stra\ss e 6, T\"{u}bingen, Germany
}

\editor{}

\maketitle

\begin{abstract}%
	Probabilistic solvers for ordinary differential equations (ODEs) provide efficient quantification of numerical uncertainty associated with simulation of dynamical systems.
	Their convergence rates have been established by a growing body of theoretical analysis. However, these algorithms suffer from numerical instability when run at high order or with small step-sizes---that is, exactly in the regime in which they achieve the highest accuracy.
	The present work proposes and examines a solution to this problem. It involves three components: accurate initialisation, a coordinate change preconditioner that makes numerical stability concerns step-size-independent, and square-root implementation.
	Using all three techniques enables numerical computation of probabilistic solutions of ODEs with algorithms of order up to 11, as demonstrated on a set of challenging test problems. The resulting rapid convergence is shown to be competitive to high-order, state-of-the-art, classical methods.
	As a consequence, a barrier between analysing probabilistic ODE solvers and applying them to interesting machine learning problems is effectively removed.
\end{abstract}

\begin{keywords}
	Probabilistic numerics, dynamical systems, ordinary differential equations, Gauss-Markov processes, state estimation
\end{keywords}

\section{Introduction}

Ordinary differential equations (ODEs) are a core concept of mechanistic modelling. Efficiently computing ODE solutions is thus important in a wide range of applications in the natural sciences.
Recently, probabilistic solvers for ODEs have emerged (see the paper by \citet{SchoberDH2014} and the references therein).
These methods are able to not only return a single point estimate that represents an approximation of an ODE solution, but they also provide uncertainty quantification calibrated to be representative of the global error \citep{schober2018, bosch2020calibrated}.
Like classical methods, they have linear complexity in the number of grid-points, and they enjoy similar convergence properties: a solver that models the ODE solution as well as its first $\nu$ derivatives ($\nu  \in \Nbb$)
can converge as fast as $h^\nu$ for step-size $h$ \citep{kersting19, tronarp20}.
However, high-order algorithms in conjunction with small step-sizes suffer from numerical instabilities: below in Section \ref{sec:improved_numerical_stability}, we show why.

Probabilistic ODE solvers cast the task of solving an ODE as a Gauss-Markov process regression problem with a non-linear observation model. This class of algorithms, at the core of probabilistic numerical methods \citep{hennig15probabilisticnumerics,cockayne19, oates19b}, builds on the large body of theory on Gaussian processes, stochastic differential equations, Bayesian filtering and smoothing as well as numerical analysis.
This blend of ideas shows in the present work: the solution to the problem of numerically stable implementation draws on concepts related to Taylor-mode automatic differentiation, Nordsieck methods, and square-root Kalman filter implementations. It solves a problem in Gauss-Markov process regression and as such, has an impact on (probabilistic) numerics and possibly every chain of computation that benefits from cheap yet effective uncertainty quantification of numerical simulation of a dynamical system. In recent years, this has turned out to be an important challenge in machine learning. .

Section \ref{sec:bayes_ode_solver} recalls the construction of probabilistic ODE solvers in the formulation as a problem of Bayesian state estimation (alternatives are provided by \citet{chkrebtii16}, \citet{conrad17}, or \citet{abdulle20}).
Section \ref{sec:improved_numerical_stability} explains and examines the tricks that are necessary to implement a high-order method.
Therein, specific parts of the algorithm are isolated. A detailed, step-by-step guide is contained in Appendix \ref{appendix:implementation_guide}.
Section \ref{sec:results} benchmarks the probabilistic ODE solver in the suggested implementation against high-order Runge-Kutta methods.
The test environments that are used throughout the paper are Lotka-Volterra, the restricted three-body problem, and a stiff version of the van der Pol system.

Matrices are capitalised ($A \in \Rbb^{d \times d}$), scalars are lower-case ($a \in \Rbb$) and matrices with a Kronecker structure are capitalised and bold-faced ($\vec{A} = A \otimes I_3 \in \Rbb^{3d\times 3d}$). Vectors that are stacks of vectors are bold-faced ($\vec{x} = (x, y, z)^\top \in \Rbb^{3d}$); generic vectors are not ($x \in \Rbb^{d}$).
Stochastic processes, e.g. the Wiener process ($w(t)$), are, with slight abuse of notation, written as functions ($x(t)$).
We use zero-indexing when describing a matrix with its elements ($A = (a_{ij})_{i,j=0, ..., N}$).
Condition numbers of matrices are computed with respect to the $\ell^2$-norm.

\section{Probabilistic ODE Solvers}
\label{sec:bayes_ode_solver}
The present work is concerned with numerically solving the $d$-dimensional, autonomous, first-order ordinary differential equation (ODE) initial value problem,
\begin{align}
	\left\{\arraycolsep=1.4pt\def\arraystretch{1.1}
	\begin{array}{rl}
		\dot x(t) & = f(x(t)), \text{ for } 0 \leq t \leq T < \infty, \\
		x(0)      & = x_0.
	\end{array}
	\right.
\end{align}
This is no loss of generality: most ODEs are autonomous, but even non-autonomous or higher-order ODEs can be translated into autonomous, first order ODEs. This would be done by writing the non-autonomous ODE as an autonomous ODE over the augmented state $\tilde x = (x(t), t)^\top$.
The restriction to autonomous ODEs simplifies the notation with respect to Jacobians of $f$, which will be required frequently throughout the subsequent exposition; especially in Section \ref{sec:initialisation}.

In the machine learning literature, the scenario in which the ODE vector field is a neural network with weights $\theta$, $f=\text{NN}_\theta(x)$, has gained traction in recent years \citep{chen2018neural,rackauckas2020universal}.
Numerically solving the initial value problem in this case corresponds to evaluating the neural network, in which the number of hidden layers corresponds to the number of evaluations that the solver takes.
Probabilistic solvers have not yet been applied to neural ODEs. As shown in the present work (and in the related study by \citet{bosch2020calibrated}), being able to faithfully integrate dynamical systems with algorithms of order $\nu = 5$ and larger implies that  approximation quality and convergence speed will not be a hindering factor for this endeavour any longer.

Other applications of ODE solvers consist of identifying mechanistic models from data. In the presence of a data set that is based on a dynamical system, determination of such a system yields not only understanding of the generative process responsible for the observations, but also compact representation of these dynamics as an ODE. \cite{kersting2020differentiable} show how probabilistic ODE solvers give rise to efficient algorithms that deal with this inverse problem.

Yet another application of computational ODE solutions lies within manifold learning: straight lines on manifolds, so-called geodesics, are computed by numerically solving a set of Euler-Lagrange ordinary differential equations. Geodesics are important for statistical analysis on manifolds, because among other things, they give rise to distance functions. \cite{hennig2014probabilistic} and \cite{arvanitidis2019fast} study the positive impact that a probabilistic ODE solver has on manifold learning.

Probabilistic ODE solvers are Gauss-Markov process regressors based on a non-linear observation model. Thus,
Section \ref{sec:prior_distribution} defines a prior distribution, Section \ref{sec:observation_model} describes the observation model and Section \ref{sec:gaussian_inference} outlines common inference strategies.
Section \ref{sec:calibration} discusses calibration and adaptive step-size selection.

\subsection{Prior Distribution}
\label{sec:prior_distribution}
This work considers Gauss-Markov priors $\vec{x}=\vec{x}(t)$ that are defined as solutions of linear, time-invariant stochastic differential equations (SDE) with Gaussian initial conditions,
\begin{subequations}
	\begin{align}
		\left\{\arraycolsep=1.4pt\def\arraystretch{1.1}
		\begin{array}{rl}
			\diff \vec{x}(t) & = \vec{F}\, \vec{x}(t) \diff t + \vec{L} \diff \vec{w}(t), \text{ for } t\geq 0, \\
			\vec{x}(0)       & \sim \Ncal(\vec{m}_0, \vec{C}_0).
		\end{array}
		\right.
	\end{align}
\end{subequations}
The vector $\vec{x}(t) = (x(t), \dot x(t), ..., x^{(\nu)}(t))^\top \in \Rbb^{d(\nu + 1)}$ models a stack of the ODE solution $x(t) \in \Rbb^d$ and its derivatives up to order $\nu \in \Nbb$.
The dispersion matrix $\vec{L}$ is, in the cases that are of interest to us, always
$\vec{L} = e_{\nu + 1} \otimes I_d \in \Rbb^{d(\nu + 1) \times d}$. $\vec{w}$ is a $d$-dimensional Wiener process with constant diffusion $\Gamma > 0$.
Choices of $\vec{F}$, $\vec{m}_0$ and $\vec{C}_0$ determine whether $\vec{x}(t)$ is, for instance, a $\nu$-times integrated Wiener process (IWP($\nu$)), a $\nu$-times integrated Ornstein-Uhlenbeck process, or a Mat\'ern process of order $\nu + 1/2$.

Let $\Tbb = \{ t_0, ..., t_N \}$ be a grid on $[0, T]$. Without loss of generality assume $t_0 = 0$ and $t_N = T$. Define the step-size $h_n = t_{n+1} - t_n$.
Restricted to $\Tbb$, there is an alternative, discretised description of the prior process such that the distribution of the continuous process $\vec{x}=\vec{x}(t)$ restricted to $\Tbb$ coincides with the distribution of the discrete process \citep{grewal2014kalman}.
Abbreviate $\vec{x}_n := \vec{x}(t_n)$; then $(\vec{x}_n)_{n = 0, ..., N}$ follows the distribution
\begin{subequations}
	\begin{align}
		\left\{\arraycolsep=1.4pt\def\arraystretch{1.1}
		\begin{array}{rl}
			\vec{x}_{n + 1} & \sim \Ncal(\vec{A}_n \vec{x}_n, \vec{Q}_n), \text{ for } n = 0, ..., N, \\
			\vec{x}_0       & \sim \Ncal(\vec{m}_0, \vec{C}_0),
		\end{array}
		\right.
	\end{align}
\end{subequations}
with matrices $\vec{A}_n \in \Rbb^{d(\nu + 1) \times d(\nu + 1)}$ and $\vec{Q}_n\in \Rbb^{d(\nu + 1) \times d(\nu + 1)}$ given by \cite[Section 6.1]{sarkka2019applied}
\begin{align}
	\vec{A}_n & := \exp(\vec{F} h_n),   \label{eq:mfd_solution1}                                                                           \\
	\vec{Q}_n & := \int_0^{h_n} \exp(\vec{F}(h_n-\tau))\vec{L} \vec{L}^\top\exp(\vec{F}^\top(h_n-\tau))\diff \tau.\label{eq:mfd_solution2}
\end{align}
In the following we will sometimes refer to $\vec{Q}_n$ as ``process noise covariance''. Both
$\vec{A}_n$ and $\vec{Q}_n$ can be computed efficiently with matrix fraction decomposition \cite[Section 6.3]{sarkka2019applied}.
For the integrated Wiener process, there exist closed form solutions to Eqs. \eqref{eq:mfd_solution1} and \eqref{eq:mfd_solution2}; we refer to Section \ref{sec:improved_numerical_stability}.

\subsection{Observation Model}
\label{sec:observation_model}
Recall the abbreviation $\vec{x}_n := \vec{x}(t_n)$.
Define the projection matrix $\vec{E}_i^\top = e_i^\top \otimes I_d\in \Rbb^{d \times d(\nu + 1)}$, where $e_i$ is the $i$th canonical basis vector in $\Rbb^{\nu + 1}$, $i=0, ..., \nu$.
Loosely speaking, $\vec{E}_i^\top$ extracts the $i$th derivative from the stack of derivatives in $\vec{x}_n$.
A probabilistic ODE solver computes a posterior distribution over $\vec{x}(t)$,
\begin{align}
	p \left(\vec{x}(t) \,\left|\,\, \vec{E}_1^\top \vec{x}_n - f(\vec{E}_0^\top \vec{x}_n) = 0\right. \right), \quad n=0, ..., N,
\end{align}
based on the prior distribution from Section \ref{sec:prior_distribution}, likelihood function
\begin{align}\label{eq:dirac_likelihood}
	\ell(\vec{x}_n) := \delta \left[\vec{E}_1^\top \vec{x}_n - f(\vec{E}_0^\top \vec{x}_n)\right], \quad n=0, ..., N,
\end{align}
and (artificial) observations at each grid point, which are all equal to zero.
This is a non-linear regression problem if $f$ is non-linear (which it usually is).
The data likelihood is a Dirac distribution composed with a non-linear function,
and inference in this model is generally intractable. 
If $\vec{x}$ is assumed to be Gaussian, approximate Gaussian filtering and smoothing yields a tractable approximation of this likelihood and hence of the posterior distribution.

\subsection{Approximate Gaussian Inference}
\label{sec:gaussian_inference}
The non-linear regression problem can be solved approximately by linearising $f$ and applying Gaussian filtering and smoothing.
Common choices are the extended Kalman filter, which linearises $f$ with a Taylor approximation, and the unscented Kalman filter, which approximates the behaviour of $f$, as it acts on Gaussian random variables, with the unscented transform \citep{sarkka2013bayesian}.

The following describes the extended Kalman filter, applied to the ODE problem \citep{tronarp19}.
Let $\vec{x}_n \sim \Ncal(\vec{m}_n, \vec{C}_n)$. The linearised observation model is
\begin{align}
	\ell(\vec{x}_n) \approx \delta\left[\vec{H}\, \vec{x}_n - \vec{b}\right].
\end{align}
$\vec{H}$ and $\vec{b}$ are derived using either a zeroth order Taylor approximation of $f(\vec{E}_0^\top \circ)$ at $\vec{m}_n$ (EK0),
\begin{align}\label{eq:linearised_model_ek0}
	\vec{H} = \vec{E}_1^\top, \quad \vec{b} = f(\vec{E}_0^\top \vec{m}_n),
\end{align}
or a first order Taylor approximation of $f(\vec{E}_0^\top \circ)$ at $\vec{m}_n$ (EK1),
\begin{align}\label{eq:linearised_model_ek1}
	\vec{H} = \vec{E}_1^\top - \nabla f(\vec{E}_0^\top\vec{m}_n) \vec{E}_0^\top, \quad \vec{b} = f(\vec{E}_0^\top \vec{m}_n) - \nabla f(\vec{E}_0^\top \vec{m}_n) \vec{E}_0^\top \vec{m}_n.
\end{align}
$\nabla f$ is the Jacobian of $f$.
If the ODE is not autonomous, the Jacobian of $f$ with respect to $x$ is used.
Both choices, EK0 and EK1, enable Gaussian filtering and smoothing algorithms; see the implementation guide in Appendix \ref{appendix:implementation_guide}.

This work, like \citet{bosch2020calibrated}, only considers the extended Kalman filter. Everything explained herein applies to the unscented Kalman filter as well, but we do not use it for reasons of computational efficiency:
Linearisation of the non-linear observation model (Eq. \eqref{eq:dirac_likelihood}) with the unscented transform requires $d(\nu + 1)$ evaluations of the ODE vector field $f$ for a single ODE solver step---one evaluation for each of the so-called \emph{sigma-points} that are used for the unscented transform.
The costs of these evaluations should be judged in comparison to the costs of evaluating the Jacobian $\nabla f$ to form $\vec{H}$ in EK1. In EK0, neither is required.

The solution to the continuous-discrete state estimation problem posed by the probabilistic ODE solver is a posterior distribution over the continuous process $\vec{x}(t)$. It can be evaluated at any time $t$, that is, in between the grid points that were used by the ODE solver to approximate the solution to the initial value problem.
In numerical analysis, this is called dense output \citep[Chapter II.6]{hairer87}. In our setting, dense output is realised by carrying out an additional, measurement-free smoothing step---and thus does not evaluate the ODE vector field $f$.
We refer to the discussion surrounding Algorithm 10.27 in the book by \citet{sarkka2019applied}.

\subsection{Calibration and Step-Size Adaptation}
\label{sec:calibration}
Efficient ODE solvers use local error control and step-size adaptation.
In probabilistic ODE solvers, the posterior covariance, which quantifies numerical uncertainty over the approximate ODE solution, benefits from post-hoc calibration of the diffusion $\Gamma$ of the driving Wiener process.
Both tasks have recently been studied by \citet{bosch2020calibrated}.
The authors present strategies for calibration of $\Gamma$ as a local quasi-maximum likelihood estimate. They extend the calibration techniques presented by \cite{schober2018} and \cite{tronarp19} by evaluating the effect of time-varying diffusion \citep{schober2018} versus time-constant diffusion \citep{tronarp19} on different variants of ODE solvers.

Uncertainty calibration can be combined efficiently with error control.
Calibrated uncertainty estimates make error estimates more meaningful and improve adaptive step-size selection \citep{bosch2020calibrated}.
In this work, we use the time-varying diffusion model together with on-the-fly calibration and the corresponding error estimate \citep{schober2018, bosch2020calibrated}.

At this point we would like to emphasise that herein, uncertainty estimates are mostly ignored, and only used to the extent that is required for error control.
We benchmark the probabilistic ODE solver \emph{as if it was a classical, deterministic method}.
Compared to related work on probabilistic numerics, this is a rather drastic point of view, and taken in order to demonstrate numerical stability and approximation quality of the algorithm. Both will be shown to be on the same level as well-established, high-order, classical numerical methods.

On top of these qualities, a probabilistic ODE solver provides uncertainty quantification in form of posterior covariances \emph{without additional costs}.
All of the run time comparisons below already include computation of posterior covariances.
Readers interested in uncertainty calibration are referred to \citet{bosch2020calibrated}.

\begin{figure}
	\centering
	\includegraphics{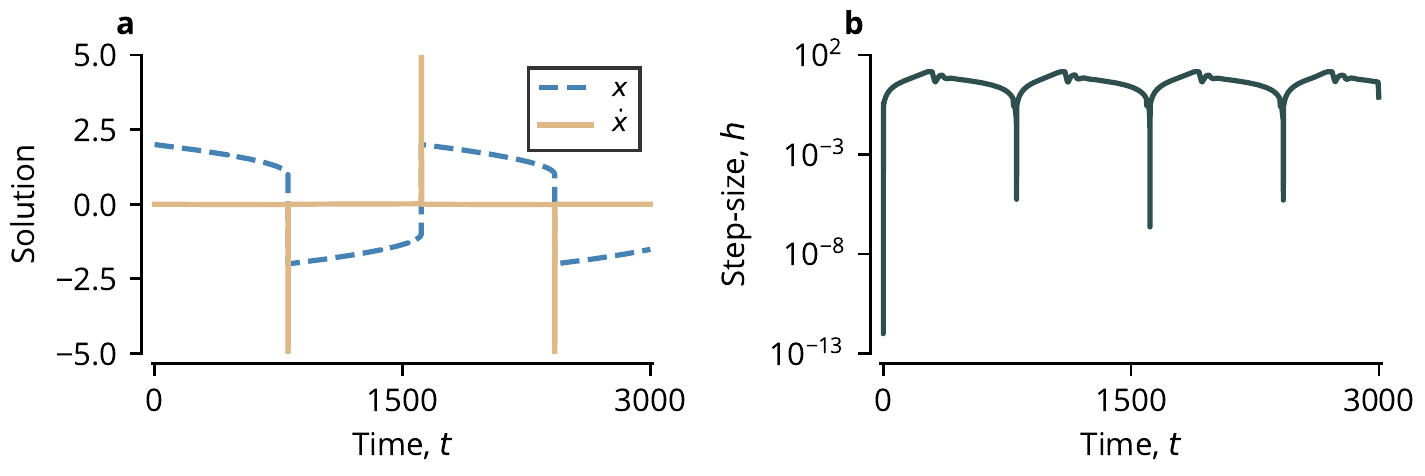}
	\caption{\textbf{Stiff van der Pol system.}
		The derivative of the solution of Eq. \eqref{eq:vanderpol} exhibits extreme spikes (a). The $y$-axis is cropped at $(-5, 5)$, but $\dot x$ takes values much larger in magnitude. The adaptive error control selects step-sizes that are extremely small (b), which shows both, the stiffness of the problem as well as the ability of the algorithm to detect, and cope with it. The first step was set to $h_0 = 0.01$, but is immediately scaled down to $\approx 10^{-12}$ by the step-size control strategy.
	}
	\label{fig:vanderpol}
\end{figure}

The EK1-solver is $A$-stable \citep[Corollary 1]{tronarp19}; a definition of $A$-stability is provided by \citet{dahlquist1963special}.
We demonstrate this stability together with the validity of the local error control scheme in Figure \ref{fig:vanderpol}, by solving the van der Pol system of ordinary differential equations \citep{guckenheimer1980dynamics},
\begin{subequations}\label{eq:vanderpol}
	\begin{align}
		\ddot x(t) & = \mu (1 - x(t)^2) \dot x(t) + x(t),
	\end{align}
\end{subequations}
from $t=0$ to $t=3000$ with initial value $(x_1(0), x_2(0)) = (2, 0)$.
We replicate the parameterisation chosen by \citet{shampine1997matlab} and set $\mu = 1000$.
With such a $\mu$, this is a stiff ordinary differential equation.
We solve this problem using EK1 with tolerance $10^{-9}$, order $\nu = 7$ and the time-varying diffusion model originally proposed by \cite{schober2018} and extended to EK1 by \citet{bosch2020calibrated}.
We only rely on standard, 64-bit floating point arithmetic as it is the default precision in NumPy arrays \citep{harris2020array}.
Without either of the implementation tricks presented below, computing the solution was impossible (especially not with order $\nu=7$).
The simulation took $\approx 9$ seconds; a reference solution with a fifth order Radau IIA solver, implemented in Scipy \citep{virtanen2020scipy}, on the exact same problem and with the same tolerance took $\approx 5$ seconds.
The van der Pol system with parameter $\mu = 10^3$ cannot be solved with EK0, likely because it does not possess the stability properties of EK1; the analysis by \cite{tronarp19} only applies to EK1 (and the unscented Kalman filter).
From Figure \ref{fig:vanderpol} it is evident how small the step-sizes must be in order for the integration to be successful. This can partly be attributed to the stiffness of the problem, but possibly also to the choice of step-control: we use proportional control, but the error estimate lends itself similarly to alternative control strategies such as PI-control \citep{gustafsson1988api}.
Future work should investigate the effect of different control strategies on adaptive step-size selection in probabilistic ODE solvers.

\section{Improved Numerical Stability}
\label{sec:improved_numerical_stability}

Exactly three components are important for the successful implementation of probabilistic ODE solvers: accurate initialisation (Section \ref{sec:initialisation}), a coordinate change in the state space model that removes instabilities for small step-sizes or high orders (Section \ref{sec:coordinate_change}), and square-root implementation of the ODE solver (Section \ref{sec:square_root_filter}).
Their implications on the overall computational complexity are discussed in Section \ref{sec:computational_complexity}.

\subsection{Accurate Initialisation}
\label{sec:initialisation}

It is important to initialise $\vec{m}_0$ and $\vec{C}_0$ as accurately as possible, for reasons of stability and approximation quality: if the initialisation is inaccurate, we report that in the best case, convergence rates do not hold and in the worst case, numerical over-/underflows happen after a few steps.
Recall that $\vec{x}$ is a stack of the ODE solution $x$ and its first $\nu$ derivatives.
Ideally, the parameters of the initial distribution, $\vec{m}_0$ and $\vec{C}_0$, are chosen as
\begin{align}
	\vec{m}_0 =
	\begin{pmatrix}
		x(0)      \\
		\dot x(0) \\
		\vdots    \\
		x^{(\nu)}(0)
	\end{pmatrix}, \quad
	\vec{C}_0 =
	\begin{pmatrix}
		0      & \cdots & 0      \\
		\vdots &        & \vdots \\
		0      & \cdots & 0
	\end{pmatrix}.
\end{align}
It is non-trivial to compute those values efficiently---in the sequel we outline one option for doing this. Alternatives are discussed in Section 3.2 in the paper by \citet{schober2018}.
\label{sec:taylor_mode}

Applying Fa\`a di Bruno's formula \citep{roman1980formula} to $y(t) := f(x(t))$ and substituting $\dot x(t) = f(x(t))$, computes higher order derivatives of $x$ at $0$. Let $0 \leq q < \nu$.
The $(q+1)$th derivative of $x$, evaluated at $t=0$ is obtained by following the recursion
\begin{align}
	\Fcal_0(x) := f(x), \quad \Fcal_{i+1}(x) := \nabla \Fcal_{i}(x) \Fcal_{i}(x), \quad i=0, ..., q -1,
\end{align}
and evaluating at zero, $x^{(q+1)}(0) = \Fcal_q(x_0)$. This approach can be implemented with automatic differentiation (AD).
Care has to be taken with the choice of AD algorithm, because if the recursive nature of the higher order derivatives is not taken into account, the complexity of AD grows exponentially with respect to $q$ \citep{kelly2020learning}.

Taylor-mode automatic differentiation is an efficient way of computing higher order derivatives of a function.
Loosely speaking, instead of tracking how to propagate directional derivatives (Jacobians), Taylor-mode AD tracks how to propagate truncated Taylor series.
Let $\hat x$ be a $\nu$th order truncated Taylor series approximation of $x$ at $t=0$, and $\hat y$ and $\hat f$ be $(\nu-1)$th order truncated Taylor series approximations of $y(t) :=f(x(t))$ and $f$, at $t=0$ and $x=x_0$ respectively,
\begin{align}
	\hat x(t) = \sum_{q=0}^{\nu} x_q t^q, \quad
	\hat y(t) = \sum_{q=0}^{\nu-1} y_q t^q, \quad
	\hat f(x) = \sum_{|\rho|=0}^{\nu-1} f_\rho (x-x_0)^\rho.
\end{align}
$\rho$ is a multi-index, because the domain of $f$ is multi-dimensional.
The coefficients $y_0, ..., y_\nu$ of $\hat y$ are computed by propagating $\hat x$ through $\hat f$. Since $x$ solves the ODE, $\dot x = y$ holds and higher order terms of $\hat x$ can be computed from lower order terms of $\hat y$, which themselves are computed from lower order derivatives of $\hat x$. More formally, the coefficients of $\hat x$ satisfy the recurrence relation $x_{q+1} = y_q / (q+1)$.
The first $\nu$ derivatives of $x$ can be read off exactly from the coefficients of $\hat x$, by definition of Taylor series.
The computational complexity of this strategy grows quadratically, sometimes only almost linearly, in the order of the approximation \citep[Chapter 13]{griewank2008evaluating}.
This is contrasted by the exponential growth in complexity in the order of the approximation of forward-mode AD.
In our Python code, we use a Taylor-mode AD implementation in JAX \citep{bradbury2020jax} based on the concept of {jets} \citep{bettencourt2019taylor}.

Computing the $\nu$th coefficient of $\hat x$, which gives the value of the $\nu$th derivative at zero, requires a $\nu$th order Taylor approximation of $f$ (everything else is computed with the iteration $x_{q+1} = y_q / (q+1)$).
Computation of an $n$th order Taylor approximation of $f: \Rbb^d \rightarrow \Rbb^d$ requires storage $\binom{n + d}{d} \approx n^d/d!$ and propagation costs $\binom{2n + d}{d}$ \citep{griewank2008evaluating}.
The complexity with Taylor-mode AD is thus significantly lower than with forward-mode AD.
For low-dimensional problems, it is almost negligible.
For high-dimensional ODEs, the costs of the initialisation need to be taken into account when choosing a high-order solver.

On a related note, efficient integration of high-dimensional ODEs with the probabilistic ODE solver is expensive, not only because of the costs of Taylor-mode AD, but also because each step of a Gaussian filter requires a sequence of matrix-matrix operations, each of which scale cubically in the dimension of the state-space.
Efficient implementation of probabilistic ODE solvers in high dimensions is a question for future research.

\subsection{Rescaled Coordinates}
\label{sec:coordinate_change}

The presentation in this section is restricted to the integrated Wiener process as a prior model. This seems to be a common choice, not only due to the intimate connection between integrated Wiener processes and polynomial splines \citep{wahba1978improper}, but also because in the whole literature on probabilistic ODE solvers, only \citet{magnani2017} and \citet{kersting2020fourier} have carried out experiments with a different prior.
It is not clear whether the following coordinate change is optimal for prior models other than integrated Wiener processes.

This section deals with ill-conditioned matrices occuring in the filtering and smoothing iterations.
Let $\vec{x}_n^F \sim \Ncal(\vec{m}_n^F, \vec{C}_n^F)$ be the filter output at the $n$th step, and let $\vec{x}_{n+1}^S \sim \Ncal(\vec{m}_{n+1}^S, \vec{C}_{n+1}^S)$ be the smoothing output at the $(n+1)$th step, that is, one step in the future.
The smoothing distribution at the $n$th step, $\vec{x}_n^S \sim \Ncal(\vec{m}_n^S, \vec{C}_n^S)$, is computed as
\begin{align}
	\vec{m}_{n+1}^- & = \vec{A}_n \vec{m}_n^F                                                                                                    \\
	\vec{C}_{n+1}^- & = \vec{A}_n \vec{C}_n^F  \vec{A}_n^\top   +  \vec{Q}_n                                       \label{eq:predict_covariance} \\
	\vec{G}_n       & = \vec{C}_{n}^F \vec{A}_n^\top (\vec{C}_{n+1}^- )^{-1}                                                                     \\
	\vec{m}_{n}^S   & = \vec{m}_{n}^F - \vec{G}_n \left[\vec{m}_{n+1}^S - \vec{m}_{n+1}^- \right]                                                \\
	\vec{C}_{n}^S   & = \vec{C}_{n}^F -  \vec{G}_n \left[ \vec{C}_{n+1}^S - \vec{C}_{n+1}^-\right] \vec{G}_n^\top,
\end{align}
and thus depends on $(\vec{C}_{n+1}^-)^{-1}$.
$ \vec{A}_n \vec{C}_n^F  \vec{A}_n^\top$ is symmetric and positive semidefinite---after all, we want $\vec{C}_n^F$ to be zero, since we want the solution to be exact.
The matrix $\vec{Q}_n$ is ill-conditioned; its representation for the IWP($\nu$) is
\begin{align}
	\vec{Q}_n = Q_n \otimes \Gamma,
	\quad Q_n = \left[ \frac{h_n^{2\nu + 1 - i - j}}{(2\nu + 1 - i - j) (\nu - i)! (\nu - j)!} \right]_{i, j=0, ..., \nu},
\end{align}
and therefore, $Q_n$ is a Hankel matrix whose entries decay rapidly from the bottom right element to the top left element.
This means that the system responds in a highly anisotropic way to step-size $h$: for some modes of the state space, the covariance increase at each step is much larger than for others, which is a problem for numerically stable computation of smoothing iterations.
In the following, we explain how to milden this ill-conditioning by means of a coordinate change in the state space model (which we equivalently refer to as a preconditioner).

Let $\vec{T} \in \Rbb^{d(\nu + 1) \times d(\nu +1)}$ be an invertible transformation matrix. The continuous-discrete system of state $\vec{x}_\text{new} := \vec{T}^{-1}\vec{x}$ is (the subscript ``new'' is omitted for readability reasons)
\begin{align}\label{eq:transformed_sde_prior}
	\left\{\arraycolsep=1.4pt\def\arraystretch{1.1}
	\begin{array}{rl}
		\diff \vec{x}(t) & = \vec{T}^{-1}\vec{F}\, \vec{T}\,\vec{x}(t) \diff t + \vec{T}^{-1}\vec{L} \diff \vec{w}(t),\text{ for }t \geq 0, \\
		\vec{x}(0)       & \sim \Ncal(\vec{T}^{-1}\vec{m}_0, \vec{T}^{-1} \vec{C}_0 \vec{T}^{-\top}).
	\end{array}
	\right.
\end{align}
The optimal choice of $\vec{T}$ will depend on step-size $h_n$. Therefore we write $\vec{T}_n := \vec{T}(h_n)$ and use a different coordinate change at each filtering/smoothing step.
The equivalent discretisation of the continuous model in Eq. \eqref{eq:transformed_sde_prior} is
\begin{align}\label{eq:transformed_equivalent_discretisation}
	\left\{\arraycolsep=1.4pt\def\arraystretch{1.1}
	\begin{array}{rl}
		\vec{x}_{n + 1} & \sim \Ncal(\vec{T}_n ^{-1}\vec{A}_n \vec{T}_n  \vec{x}_n, \vec{T}_n ^{-1}\vec{Q}_n \vec{T}_n ^{-\top}), \text{ for } n = 0, ..., N, \\
		\vec{x}_0       & \sim \Ncal(\vec{T}_0^{-1}\vec{m}_0,\vec{T}_0^{-1} \vec{C}_0\vec{T}_0^{-\top}).
	\end{array}
	\right.
\end{align}
The measurement model changes as well. It now reads
\begin{align}\label{eq:transformed_measurement_model}
	\ell(\vec{x}_n) 
	\approx \delta \left[\,\overline{\vec{H}}\,\vec{T}_n\, \vec{x}_n - \overline{\vec{b}} \,\right],
\end{align}
which assumes that the state $\vec{x}_n$ ``lives in the preconditioned space''. $\overline{\vec{H}}$ and $\overline{\vec{b}}$ are derived by linearising $f(\vec{E}_0^\top\vec{T}_n\circ)$ with a zeroth or first order Taylor approximation at $\vec{m}_n$ (recall Eqs. \eqref{eq:linearised_model_ek0} and \eqref{eq:linearised_model_ek1}).
The filtering and smoothing iterations are changed accordingly. A detailed implementation guide is in Appendix \ref{appendix:implementation_guide}.

Next, we propose such a coordinate change. If in Eqs. \eqref{eq:transformed_equivalent_discretisation} and \eqref{eq:transformed_measurement_model}, $\vec{T}_n$ is chosen as
\begin{align}\label{eq:transformation_matrix}
	\vec{T}_n := T_n \otimes I_d, \quad T_n := \sqrt{h_n}\diag\left(\frac{h_n^\nu}{\nu!}, \frac{h_n^{\nu-1}}{(\nu-1)!}, ..., h_n, 1\right),
\end{align}
the dependency of $\vec{A}_n$ and $\vec{Q}_n$ on $h_n$ is removed, because those two matrices can be factorised as
\begin{align} \label{eq:factorisation_a_q}
	\vec{A}_n = {T}_n \,\overline{{A}} \,{T}_n^{-1} \otimes I_d, \quad \vec{Q}_n = {T}_n \,\overline{{Q}} \,{T}_n^\top \otimes \Gamma.
\end{align}
Applying the coordinate change to $\vec{A}_n$ and $\vec{Q}_n$ leaves only $\overline{\vec{A}} = \overline{A} \otimes I_d$ and $\overline{\vec{Q}} = \overline{Q} \otimes \Gamma$, because $\vec{T}_n^{-1}$ and $\vec{T}_n$ cancel each other out (compare Eq. \eqref{eq:transformed_equivalent_discretisation} to Eq. \eqref{eq:factorisation_a_q}).
The upper triangular matrix $\overline{{A}}$ as well as the Hankel matrix $\overline{{Q}}$ are available in closed form. They are
\begin{align}
	\overline{A} := \left[ \binom{\nu - i}{\nu - j}\right]_{i,j=0, ..., \nu} \quad \text{and}\quad
	\overline{Q} :=  \left[\frac{1}{2\nu + 1-i - j}\right]_{i,j=0, ..., \nu},
\end{align}
where the elements of $\overline{A}$ are binomial coefficients.
Removing the $h_n$-dependency from the discretisation is crucial, because (i) the elements---and hence, the condition number---of the process noise covariance are independent of the step-size and (ii) this transformation can be computed in closed form and applied to an $d(\nu + 1) \times d(\nu + 1)$ matrix in complexity $\Ocal(d^2(\nu + 1)^2)$, which is neglibile if viewed in the context of the matrix-matrix operations in each ODE solver step.
Cheap application of the preconditioner is of utmost importance, because ODE solver implementations need to be fast. We refer to Section \ref{sec:computational_complexity} for a more thorough complexity analysis.

This transformation implies that, although we store the values $(x, \dot x, ..., x^{(\nu)})$, we work in the rescaled coordinates $(h^{-\nu} x \, \nu!, ..., x^{(\nu)})$.
Even though the coordinate systems are different, this is conceptually related to and, in fact, inspired by the Nordsieck representation of a vector $(x, ..., h^\nu x^{(\nu)}/\nu!)$ \citep{nordsieck1962numerical}.
Such a representation, which we refer to as Nordsieck coordinates, was used by \citet{schober2018} to show that the zeroth order linearisation filter (EK0) is a multi-step method with time-varying weights.
It has therefore been proven useful to analyse the probabilistic ODE solver. The evaluation below will show that a variant of this change additionally solves problems of numerical stability.

\begin{figure}
	\centering
	\includegraphics{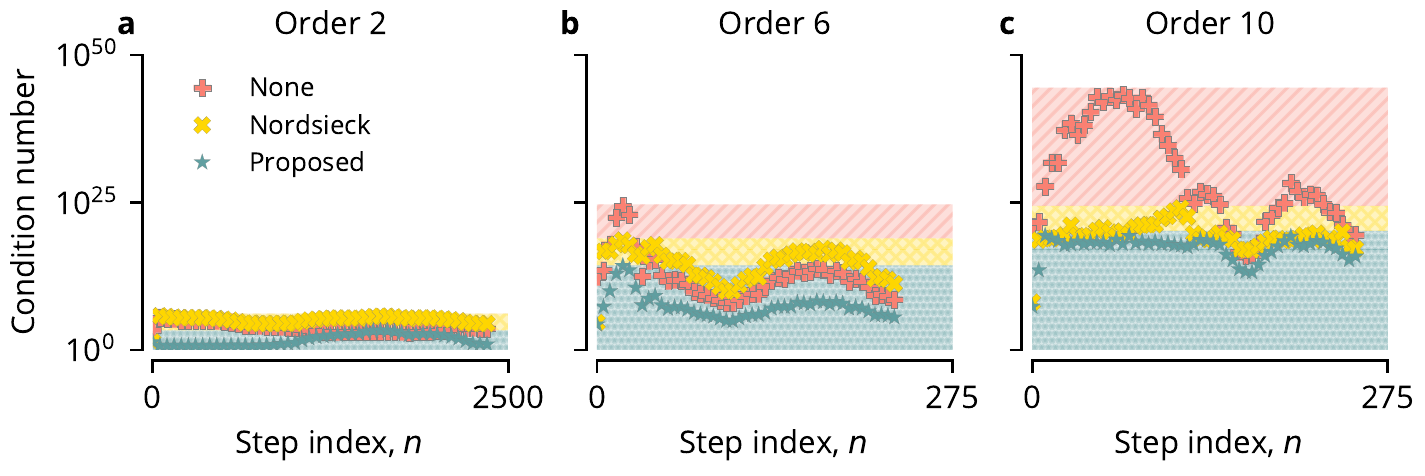}
	\caption{\textbf{Conditioning of the Cholesky factors of the predictive covariances on a test-problem.} Condition numbers of the Cholesky factors of the predictive covariances ``as seen by the solver''; that is, an ODE solution is computed with adaptive step-sizes and tolerance $10^{-4}$ using the proposed coordinate change and at each step, the condition number of ``what would have been predicted in Nordsieck/original coordinates'' is computed.
		Evaluated for orders $\nu=2$ (a), $\nu=6$ (b), and $\nu=10$ (c).
		Comparison of no preconditioning (red plus), Nordsieck coordinates (yellow cross) and the proposed change (blue star).
		The maximum condition number on this interval decides whether the smoothing iteration fails or succeeds. The discrepancy between the maximal condition numbers is shaded in red (None vs. Nordsieck), and yellow (Nordsieck vs. Proposed). The range between 0 and the maximum condition number of the proposed change is shaded blue.
		The underlying ODE is the Lotka-Volterra model in the parameterisation from the experiments below.  }
	\label{fig:cond_numbers}
\end{figure}

The proposed change improves the condition number of the process noise covariance $\vec{Q}_n$ and the predictive covariance $\vec{C}_{n+1}^-$ more than Nordsieck coordinates do; see Table \ref{table:taylor_nordsieck} (and Figure \ref{fig:cond_numbers}; more on this below).
\begin{table}
	\centering
	\begin{tabular}{| c | c c c | c c c | c c c | }
		\hline
		\multirow{2}{*}{Order, $\nu$}
		   & \multicolumn{3}{|c|}{$\log_{10}[\cond(Q)]$} & \multicolumn{3}{|c|}{$\log_{10}[\rho]$} & \multicolumn{3}{|c|}{$\log_{10}[\min(\lambda_i(Q))]$}                                                                      \\\cline{2-10}
		   & Prop.                                       & Nord.                                   & None                                                  & Prop.     & Nord.   & None  & Prop.      & Nord.      & None       \\
		\hline
		1  & \bf 1.3                                     & \bf 1.3                                 & 9.1                                                   & \bf  0.5  & \bf 0.5 & 8.5   & \bf  -1.2  & -5.2       & -13.1      \\
		3  & \bf  4.2                                    & 4.3                                     & 28.9                                                  & \bf  0.8  & 1.3     & 26.4  & \bf -4.0   & -9.1       & \Lightning \\
		5  & \bf 7.2                                     & 7.6                                     & 43.7                                                  & \bf 1.0   & 2.3     & 45.2  & \bf  -7.0  & -14.1      & \Lightning \\
		7  & \bf 10.2                                    & 11.0                                    & 57.3                                                  & \bf   1.2 & 3.4     & 64.6  & \bf  -10.0 & -19.8      & \Lightning \\
		9  & \bf 13.2                                    & 14.5                                    & 68.5                                                  & \bf  1.3  & 4.5     & 84.4  & \bf -13.0  & -25.9      & \Lightning \\
		11 & \bf 16.2                                    & 17.4                                    & 79.9                                                  & \bf  1.4  & 5.6     & 104.6 & \bf -16.0  & \Lightning & \Lightning \\
		\hline
	\end{tabular}
	\caption{\textbf{Conditioning of the preconditioned process noise covariance.} Evaluation of the condition number of the process noise covariance $\vec{{Q}}_h$ after preconditioning (left column), the ratio of largest element and the smallest element $\rho = \max_{ij}(\vec{Q}_h)_{ij} / \min_{ij}(\vec{Q}_h)_{ij}$ of this matrix (middle column; all elements are positive), and the smallest eigenvalue (right column). Evaluated are the proposed coordinated change (Prop.), Nordsieck coordinates (Nord.) and no preconditioning (None). The latter two are computed with $h=10^{-4}$, which we argue to be a realistic scenario for an ODE solver; the former is step-size independent. Values are displayed in $\log_{10}$ basis and rounded to a single decimal. The ``best'' values (i.e.\ smallest in magnitude) are bold-faced---they all use the proposed coordinate change. NaN's are marked with a lightning (\Lightning), which occurs if the logarithm of a negative number is taken---numerically, the matrix is not positive definite anymore.
	}
	\label{table:taylor_nordsieck}
\end{table}
Without preconditioning, numerical instability is severe for $\nu > 1$; the competition between Nordsieck coordinates and the proposed change is close.

An evaluation of the effect of different coordinate changes on the condition numbers of the predictive covariances $(\vec{C}_{n+1}^-)_{n=0, ..., N}$ is displayed in Figure \ref{fig:cond_numbers}.
From this experiment, two conclusions are evident: (i) non-preconditioned systems have large condition numbers; (ii) Nordsieck coordinates and the proposed preconditioner both remedy this problem, though Nordsieck coordinates perform worse than the transformation from Eq. \eqref{eq:transformation_matrix}.
This is aligned with the information in Table \ref{table:taylor_nordsieck}.
In summary: the ratio of the elements and eigenvalues in the preconditioned process noise covariance as well as the condition number of the predictive covariance speak in favour of using the proposed transformation $\vec{T}_n$ over Nordsieck coordinates, though both are better than no preconditioning.

\subsection{Square-Root Kalman Filter}
\label{sec:square_root_filter}

Even with good initialisation and rescaled coordinates, numerical instability affects the implementation negatively.
The reason is that the covariance matrices may have some negative eigenvalues due to round-off errors and finite precision arithmetic.
Classical ODE solvers do not have this problem, because they do not provide the same uncertainty quantification.
Probabilistic ODE solvers propagate uncertainty estimates in the form of  covariance matrices $\vec{C}_0$, ..., $\vec{C}_N$, which need to be statistically valid, i.e.\ symmetric and positive \mbox{(semi-)}definite, even for small steps and high orders.

Symmetric, positive definite matrices allow Cholesky factorisations.
Symmetric, positive semidefinite matrices do not allow Cholesky factorisations but matrix square-roots (e.g. computed with an LDL decomposition, a close relative of a Cholesky decomposition).
Matrices that need to be inverted will be guaranteed to be positive definite (see Appendix \ref{appendix:implementation_guide}); for intermediate calculations, any matrix square-root is sufficient.

If the filtering algorithm tracks matrix square-roots instead of full covariance matrices and applies all subsequent linear algebra operations to these square-roots only, positive (semi-)definiteness and symmetry are preserved throughout the entire iteration.
This is the square-root Kalman filter. According to \citet[p. 18]{grewal2014kalman}, it dates back to \cite{potter1963statistical}, and is known to solve numerical instability issues \citep[Chapter 7]{grewal2014kalman}.

Let $\vec{C}_n = \vec{L}_C \vec{L}_C^\top$ be any matrix square-root factorisation of the covariance $\vec{C}_n$. The subscript $n$ is omitted in $\vec{L}_C$ for readability reasons.
Similarly, let $\vec{Q}_n = \vec{L}_Q \vec{L}_Q^\top$. Then, the right-hand side of  Equation \eqref{eq:predict_covariance}, which computes the predicted covariance, 
is the product of two $d(\nu + 1) \times 2d(\nu + 1)$ matrices
\begin{align}
	\vec{C}_{n+1}^- =
	\begin{pmatrix}
		\vec{A}_n \vec{L}_C & \vec{L}_Q
	\end{pmatrix}
	\begin{pmatrix}
		\vec{L}_C^\top \vec{A}_n^\top \\
		\vec{L}_Q^\top
	\end{pmatrix}.
\end{align}
The QR decomposition factorises $(\vec{A}_n \vec{L}_C, \vec{L}_Q)^\top$ as
\begin{align}
	\begin{pmatrix}
		\vec{L}_C^\top \vec{A}_n^\top \\
		\vec{L}_Q^\top
	\end{pmatrix}
	=
	\vec{X}     \begin{pmatrix}
		\vec{R} \\
		\vec{0}
	\end{pmatrix},
\end{align}
for an orthogonal matrix $\vec{X} \in \Rbb^{d(\nu + 1) \times d(\nu + 1)}$ (the variable name ``$\vec{Q}$'' is already assigned to the process noise covariance) and an upper triangular matrix that stacks an upper triangular matrix $\vec{R}$ on top of zeros.
$\vec{R}^\top$ is the Cholesky factor of $\vec{C}_{n+1}^-$,
\begin{align}
	\vec{C}_{n+1}^- =
	\begin{pmatrix}
		\vec{A}_n \vec{L}_C & \vec{L}_Q
	\end{pmatrix}
	\begin{pmatrix}
		\vec{L}_C^\top \vec{A}_n^\top \\
		\vec{L}_Q^\top
	\end{pmatrix}
	=
	\begin{pmatrix}
		\vec{R}^\top & 0
	\end{pmatrix} \vec{X}^\top \vec{X}
	\begin{pmatrix}
		\vec{R} \\ 0
	\end{pmatrix} = \vec{R}^\top \vec{R}.
\end{align}
The matrix $\vec{R}$ is unique up to multiplication with the matrix $\diag(\pm 1, ..., \pm 1)$.
Multiplying such a matrix to $\vec{R}$ from the left and to $\vec{X}$ from the right ensures that the diagonal of $\vec{R}$ is always positive, which makes it a valid Cholesky factor, while preserving a valid QR decomposition, because the orthogonal matrix remains orthogonal.

The same trick can be applied to computing the Cholesky factor of the product of matrices $\vec{H} \vec{L} \vec{L}^\top\vec{H}^\top$ where $\vec{H} \in \Rbb^{d(\nu + 1)}$ is not quadratic.
This is important for the update step of the EK0 or EK1.
We refer to Appendix \ref{appendix:implementation_guide} below.

\subsection{Computational Complexity}
\label{sec:computational_complexity}
Assembly of $\vec{A}_n$ and $\vec{Q}_n$ at each step is replaced with pre-computation of $\overline{\vec{A}}$ and $\overline{\vec{Q}}$, which saves valuable computing time.
$\vec{T}_n$ is diagonal and therefore, applying (and undoing) the preconditioner is cheap.

Inversion of covariance matrices, which is required for assembly of Kalman- and smoothing-gain, respectively, is expedited because of the readily computed Cholesky decomposition.
The complexity of computing a QR decomposition of a matrix $M \in \Rbb^{m \times n}$, $m \geq n$, is $\Ocal(m n^2)$ \citep[Table C.2]{higham2008functions}, and thus in the same class as matrix-matrix multiplication.
The latter is a prominent operation in the Kalman filter, so the asymptotical complexity of a single step of the ODE solver remains unaffected by the choice of square-root implementation over the ``classical'' implementation.
In practice, the QR decomposition slightly increases the run time of the algorithm.
Future work may consider implementing an efficient QR decomposition that exploits the sparsity pattern in e.g. $(\overline{\vec{A}} \vec{L}_n, \vec{L}_Q)^\top$, where the bottom half is triangular.
In light of gaining numerical stability to the point where previously unfeasible algorithms can be implemented robustly, a small increase in computing time seems affordable.

\subsection{Summary}

This concludes the list of implementation tricks that are necessary to implement high-order probabilistic ODE solvers. Section \ref{sec:taylor_mode} introduced accurate initialisation with Taylor-mode automatic differentiation, which is an automatic differentiation framework ``tailored'' to propagation of truncated Taylor series; Section \ref{sec:coordinate_change} explained that with a small twist on classical Nordsieck vector coordinate systems, numerical stability concerns in an ODE solver are step-size independent; Section \ref{sec:square_root_filter} explained how to change the implementation of the filter step in order to track only the matrix square-roots of covariance matrices, which ensures positive semidefiniteness and symmetry throughout the iteration.
A detailed, step-by-step implementation guide using all three proposed modifications is contained in Appendix \ref{appendix:implementation_guide}.
Next, in Section \ref{sec:results}, the effectiveness of the new scheme will be demonstrated.

\section{Results}
\label{sec:results}

This section investigates how the proposed changes affect computation of ODE solutions with high-order probabilistic solvers.
At first, we show work-precision diagrams for the Lotka-Volterra system. This is a simple ODE problem, which can be computed to high precision with most ODE solvers. We hope to see rapid convergence for high-order methods, for both EK0 and EK1.
We compare the probabilistic EK1 solver against Scipy implementations of Runge-Kutta methods.
Afterwards, we repeat the same benchmarks on the three-body-problem, which is a tougher ODE to solve than Lotka-Volterra.

We evaluate the final time error, which measures the discrepancy between the approximate ODE solution and a reference ODE solution at the final time point $t=T$.
The final time error is the same error measure for both filtering and smoothing implementations; thus, by considering this error we can relate to the convergence rates by \cite{kersting19} who consider only filtering algorithms.
We also use the root mean-square error (RMSE) on an equidistant grid $\{0, h, ..., N/h =T\}$ with resolution $h=10^{-2}$. This grid is different to the grid that is used for computation of the solution (which uses adaptive step-size selection).
The RMSE is an approximation of the $L^2$ distance, and is chosen to show off numerical stability of the smoothing steps, which are required to compute dense output, i.e.\ to evaluate the approximate ODE solution between two grid points (recall Section \ref{sec:gaussian_inference}). If both smoothing and dense output are numerically stable \emph{and} convergence rates of roughly $h^\nu$ are matched, the implementation is sufficiently stable.

\subsection{Lotka-Volterra}

We begin the experiments by numerically integrating the Lotka-Volterra predator-prey model \citep{lotka1978growth},
\begin{subequations}
	\begin{align}
		\dot x_1(t) & = 0.5 x_1(t) - 0.05 x_1(t) x_2(t)    \\
		\dot x_2(t) & = -0.05  x_2(t) + 0.5 x_1(t) x_2(t),
	\end{align}
\end{subequations}
from  $t_0 = 0$ to $T = 20$, initialised at $x_1(0) = x_2(0) = 20$. $x_1$ is the number of prey and $x_2$ is the number of predators. The coefficients describe the interaction of the two species.

\begin{figure}
	\centering
	\includegraphics{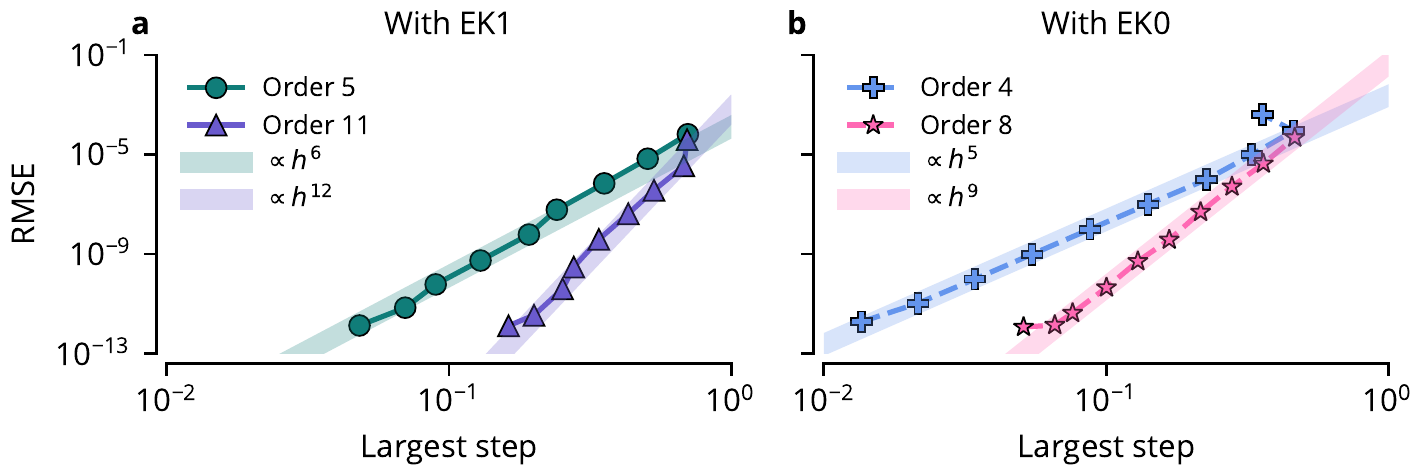}
	\caption{\textbf{EK0 and EK1 on Lotka-Volterra.}
		Convergence rates for the probabilistic ODE solver using EK1 and orders $\nu=5,11$ (a); and using EK0 and orders $\nu = 4, 8$. Convergence rates of at least $h^\nu$ hold, even for $\nu = 11$. The curves taper off at around $10^{-12}$ because the accuracy of the reference solution is reached.
	}
	\label{fig:lotkavolterra_convergence}
\end{figure}

The reference solution is computed with Scipy's RK45 and tolerance $\epsilon_{\text{abs}} = 10^{-13} =\epsilon_{\text{rel}} = 10^{-13}$.
Convergence rates for EK0 and EK1 are shown in Figure \ref{fig:lotkavolterra_convergence}, where the RMSE is plotted against the largest step, also known as fill distance.
High-order convergence rates are visible for both EK0 and EK1 and all depicted orders---even for $\nu = 11$.

Strictly speaking, this demonstration of EK0 convergence does not fall into the setting of the convergence rates analysed by \citet{kersting19}, because we use a time-varying diffusion model.
Nevertheless, the visible convergence rates of at least $h^{\nu}$ in Figure \ref{fig:lotkavolterra_convergence} confirm the numerical stability of the implementations and strengthen the conjecture by \citet{kersting19} about the generalisability of their convergence rates from $\nu = 1$ to $\nu \gg 1$.
EK1 is neither part of the analysis by \citet{kersting19}, which describe zeroth order linearisation (EK0), nor part of the theory by \citet{tronarp20}, which are concerned with the MAP estimate (which can be computed by iterated extended Kalman smoothing).
Though, arguably, one might speculate that similar convergence rates hold for EK1.

\begin{figure}
	\centering
	\includegraphics{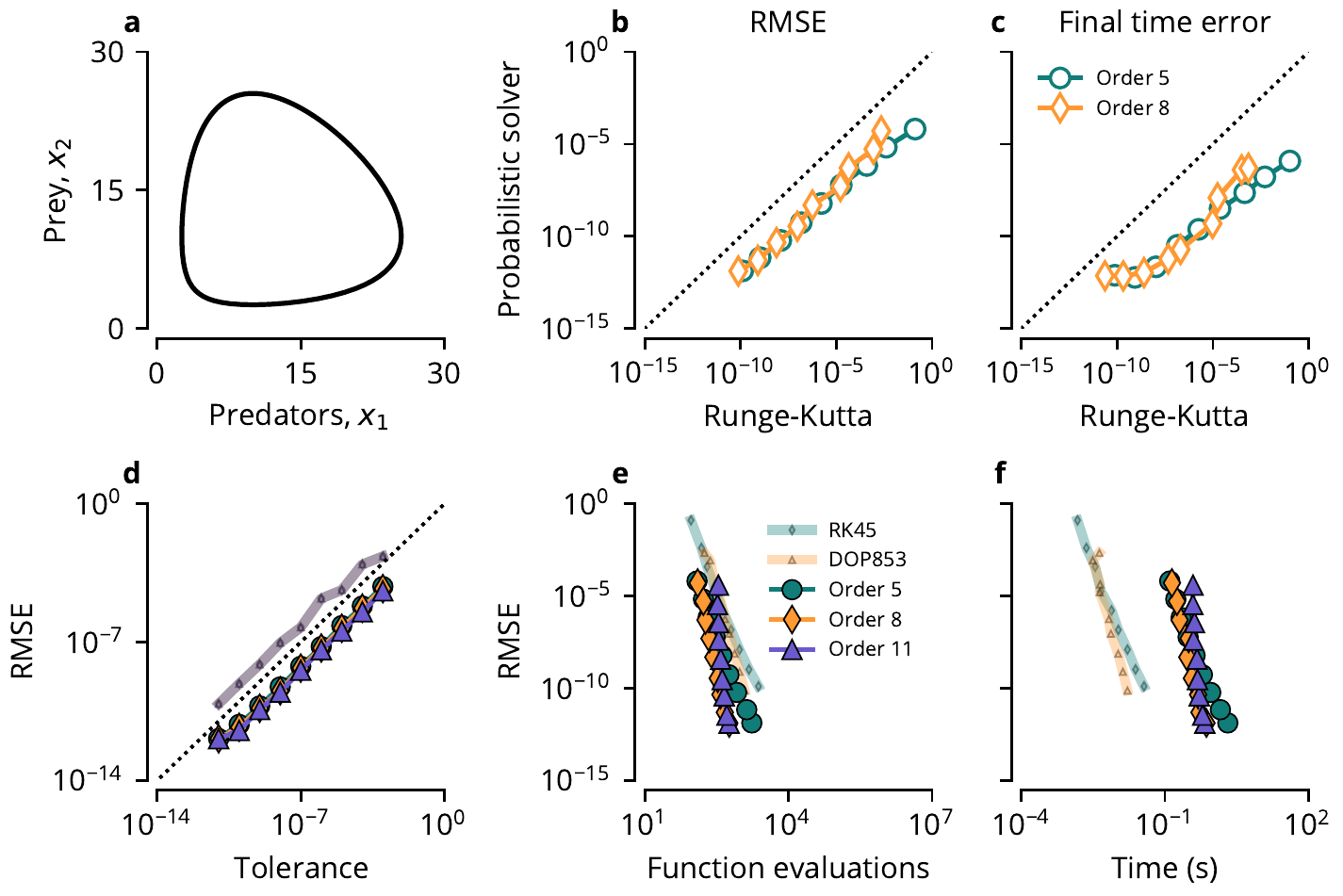}
	\caption{\textbf{Detailed results on Lotka-Volterra.}
		The reference solution is periodic (a). In terms of both RMSE and error at final time, the probabilistic EK1-solver converges to the reference solution as fast as Scipy's Runge-Kutta methods (b, c)
		For error $\approx 10^{-12}$, the curves taper off because the accuracy of the reference solution is reached (c). The RMSE responds well to a user-specified tolerance (d). In terms of function evaluations and time, \emph{the solver converges fast, especially for high orders (e, f).}
	}
	\label{fig:results_lotkavolterra}
\end{figure}

Next, we evaluate the performance of EK1 against Runge-Kutta implementations in Scipy.
The results are depicted in Figure \ref{fig:results_lotkavolterra} and confirm the efficiency of the scheme.
The probabilistic solver, based on EK1, exhibits a convergence rate of order 12 for $\nu = 11$---the work-precision curve of $\nu = 11$ in Figure \ref{fig:results_lotkavolterra} is almost vertical---which is beyond the capabilities of Scipy's ODE solver suite.
Its runtime is proportional to Scipy's Runge-Kutta methods; more specifically, it is longer with factor $\sim 10$. Given that it requires a sequence of matrix-matrix operations, and Runge-Kutta methods do not, this is a positive result.

\subsection{Three-Body}

In the second example, we try the numerical solution of the restricted three-body problem as described by \cite{hairer87}. It models the trajectory of a body in the gravitational system between the moon and earth.
Let $\mu_1 = 0.012277471$ be the standardised moon-mass and $\mu_2 := 1 - \mu_1$. The solution of the ODE
\begin{subequations}
	\begin{align}
		\ddot x_1(t) & = x_1(t) + 2 \dot x_2(t) - \mu_2 \frac{x_1(t) + \mu_1}{D_1(t)} - \mu_1 \frac{x_1(t) - \mu_2}{D_2(t)} \\
		\ddot x_2(t) & = x_2 - 2 \dot x_1(t) - \mu_2 \frac{x_2(t)}{D_1(t)} - \mu_1 \frac{x_2(t)}{D_2(t)}                    \\
		D_1(t)       & = ((x_1(t) + \mu_1)^2 + x_2(t)^2)^{3/2}                                                              \\
		D_2(t)       & = ((x_1(t) - \mu_2)^2 + x_2(t)^2)^{3/2}
	\end{align}
\end{subequations}
is periodic on  $t \in [t_0, T] = [0, 17.0652165601579625588917206249]$, if initialised with $x_1(0) = 0.994$, $x_2(0) = 0$, $\dot x_1(0) = 0$, and  $\dot x_2(0) = -2.00158510637908252240537862224$.
Every decimal in $\dot x_2(0)$ and $\mu_1$ respectively matters---if ignored, the solution is not periodic.
This problem, although classified by \cite{hairer87} as non-stiff, is a much more challenging simulation than Lotka-Volterra. $f$ has two singularities (at $(x_1, x_2)=(-\mu_1, 0)$ and at $(x_1, x_2) = (\mu_2, 0)$ respectively), and close to those singularities, much smaller steps are required to achieve given accuracy, than far away from the singularities.

\begin{figure}
	\centering
	\includegraphics{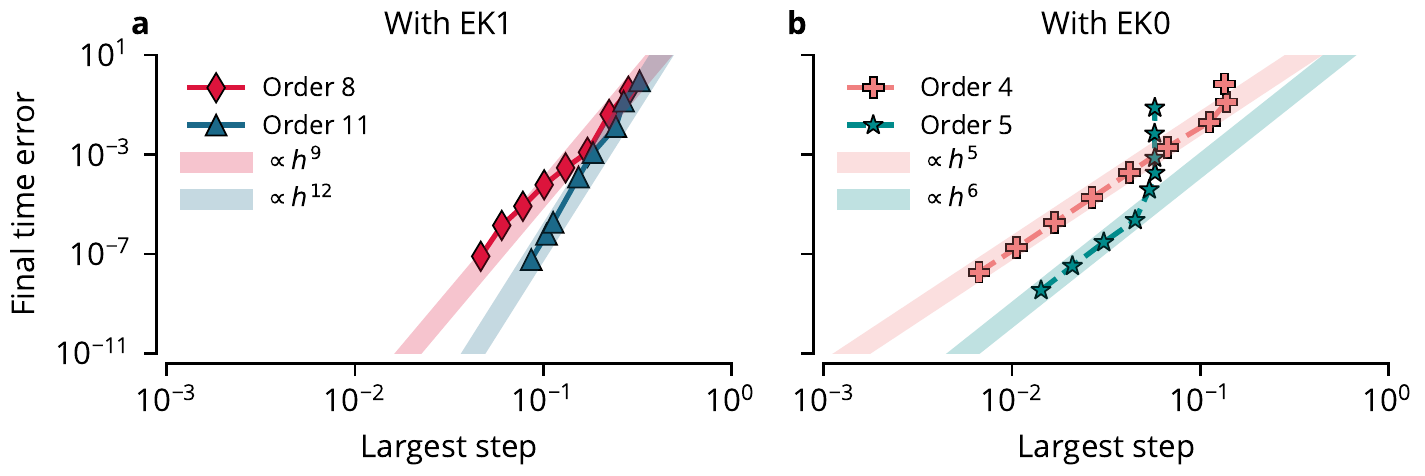}
	\caption{\textbf{EK0 and EK1 on the three-body problem.}
		Convergence rates for the probabilistic ODE solver using EK1  and orders $\nu=8, 11$ (a); and using EK0 and orders $\nu = 4, 5$ (b). Again, convergence rates of at least $h^\nu$ hold, even for $\nu = 11$.
		This time, we evaluate the error at the final time point $t=T$, choosing $T=25.5978248402$ (which is 150$\%$ of the period of the problem).
		Both solvers exhibit reliable, fast convergence for all depicted orders.
	}
	\label{fig:threebody_convergence}
\end{figure}

We compute a reference solution with LSODA \citep{hindmarsh2005lsoda},  and tolerance $\epsilon_{\text{abs}} = 10^{-12} =\epsilon_{\text{rel}} = 10^{-12}$.
LSODA is chosen, because (i) it has automatic stiffness detection and switching, which copes well with the challenges the three-body problem poses, and (ii) because it is neither RK45 nor DOP853 and thus does not bias the work-precision diagrams.\footnote{On Lotka-Volterra we were not concerned by this bias, because of the simplicity of the problem and the fact that almost every solver performs well.}
Even on this comparably tough problem, high polynomial convergence rates seem to hold; see Figure \ref{fig:threebody_convergence}.
EK1 exhibits stable $h^\nu$ convergence even for $\nu = 11$. For EK0, the same rates are visible for $\nu = 4, 5$. Higher orders of EK0 still converged, but adaptive step-size selection was less efficient than for $\nu = 5$.
For both solvers, we see \emph{faster} convergence than $h^\nu$; like in Figure \ref{fig:lotkavolterra_convergence} (Lotka-Volterra), we observe rate $h^{\nu + 1}$. We do not investigate this faster-than-expected convergence further in this work.

\begin{figure}
	\centering
	\includegraphics{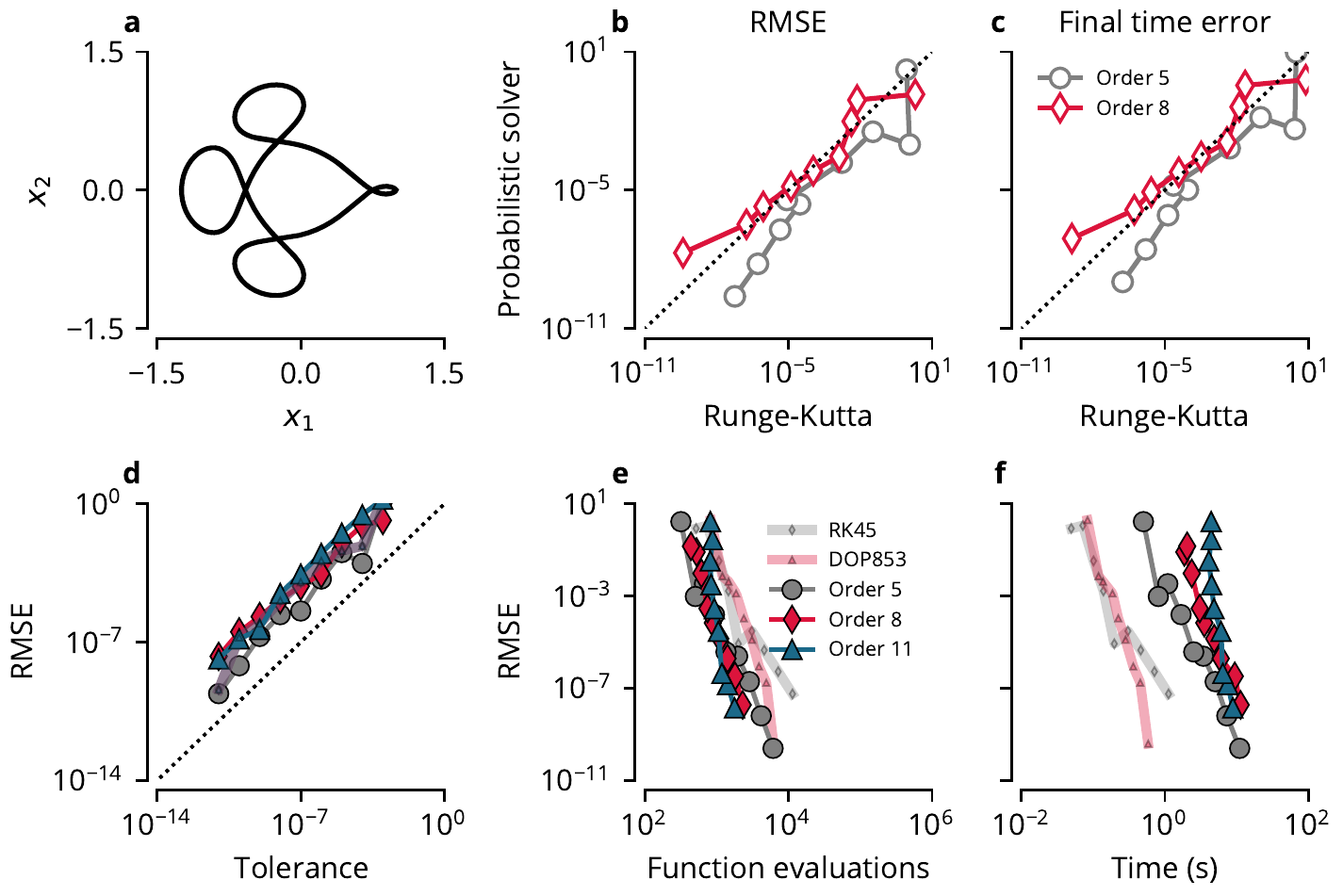}
	\caption{\textbf{Detailed results on the three-body problem.}
		A successful solution is periodic (a). The probabilistic  EK1-solvers of orders $\nu = 5$ and $\nu=8$ match the performance of Scipy implementations of Runge-Kutta methods of equal order  exactly (RK45, DOP853; b, c), which is both measured in the root-mean-square error  (RMSE; b) and in the error at final time $t=T$ (c). The RMSE improves proportionally to the user-specified tolerance (d), though it appears to lack behind with factor $\sim 10$. The same applies to Runge-Kutta solvers, whose lines are hardly visible in (d), because they have so much overlap with the markers of the probabilistic solvers.
		Solvers of order $\nu=5, 8, 11$ converge fast (e, f). The runtime is proportional to the Scipy implementations of Runge-Kutta methods, up to factor $\sim 10$ (f).
	}
	\label{fig:results_threebody}
\end{figure}

More detailed simulation results are depicted in Figure \ref{fig:results_threebody}. We compare the runtime and accuracy of the probabilistic EK1-solver against reference Runge-Kutta implementations in Scipy.
The performance of probabilistic solvers seems to be comparable to well-established, non-probabilistic solvers.

\subsection{Summary of the Experiments}

Implementation was numerically stable for both EK0 and EK1 and all orders $1 \leq \nu  \leq 11$ in a way that (i) convergence rates are visible in work-precision diagrams even for order $\nu = 11$ on the three-body problem, and (ii) convergence is at least as fast as for Runge-Kutta methods of comparable order.

\begin{wraptable}{r}{0.45\textwidth}
	\centering
	\begin{tabular}{ |c|c|c| }
		\hline
		    & Stiff               & Non-Stiff            \\
		\hline
		EK0 & Do not use it.      & $4 \leq \nu \leq 8$  \\
		EK1 & $4 \leq \nu \leq 7$ & $4 \leq \nu \leq 11$ \\
		\hline
	\end{tabular}
	\caption{Which orders $\nu$ can be recommended based on the present experiments?}
	\label{table:which_order}
\end{wraptable}
It is difficult to recommend an optimal choice of $\nu$, because this decision will likely be problem-specific. The following is our experience with numerical simulation of the above ODEs.
Orders $\nu < 4$ converged slowly and in all scenarios, $\nu \geq 4$ was feasible.
EK0 performed best with $\nu \leq 9$, so a good range for EK0 appears to be $4 \leq \nu \leq 8$.
For a more involved problem, like the simulation of the three-body dynamics, $4 \leq \nu \leq 5$ was most efficient.
EK1 showed rapid convergence on both non-stiff test problems for orders up to 11.
Since the dimension of the state space is $d(\nu + 1)$, choosing a high order comes at the price of computational complexity.
It seems that high orders go well with low tolerances, i.e.\ high accuracy, but this conjecture requires further research.
On the stiff van der Pol problem (Section \ref{sec:calibration}), orders $\nu \geq 7$ were unstable and did not converge, and orders $\nu \leq 4$ converged slowly. EK0 was impossible, likely due to a lack of $A$-stability.
We summarise these findings in Table \ref{table:which_order}.

While the sole focus of the present investigation was showing that even if benchmarked ``as a classical method'' the probabilistic ODE solver is competitive to high-order Runge-Kutta methods, at this point we would like to recall that with the probabilistic algorithm, uncertainty quantification in the form of a posterior covariance comes for free---that is, computation of this quantity is \emph{already} contained in the runtime analysis detailed above.
The value of this uncertainty quantification for solving inverse problems has been demonstrated by \citet{kersting2020differentiable}.

\section{Discussion}
\label{sec:discussion}

The presented transformations evidently allow computation of ODE solutions with a probabilistic ODE filter/smoother and orders $\nu \gg 1$, which to the best of the author's knowledge has not been possible before.
Limits are given only by numerical (i.e.\ asymptotic, that is, $A$-, $B$- or $L$-) stability of the algorithm and computational efficiency for high-dimensional or stiff problems. These are questions that are not only important for the probabilistic solver, but need theoretical analysis for many other methods, too.
The presented guide enables empirical research on answering these questions.

The experiments show that Taylor-mode AD, a coordinate change in the state space, and square-root implementation of the filter are an improved implementation in terms of numerical stability, even over the Nordsieck-transformation that is mentioned by \citet{schober2018}, a variant of which has been used in \texttt{ProbNum}, a collection of probabilistic numerical algorithms in Python.
The implementation will be made available in \texttt{ProbNum}:
\begin{align*}
	\begin{split}
		~~~~~~~~~~~~~~~~~~~~~\includegraphics[height=0.75cm]{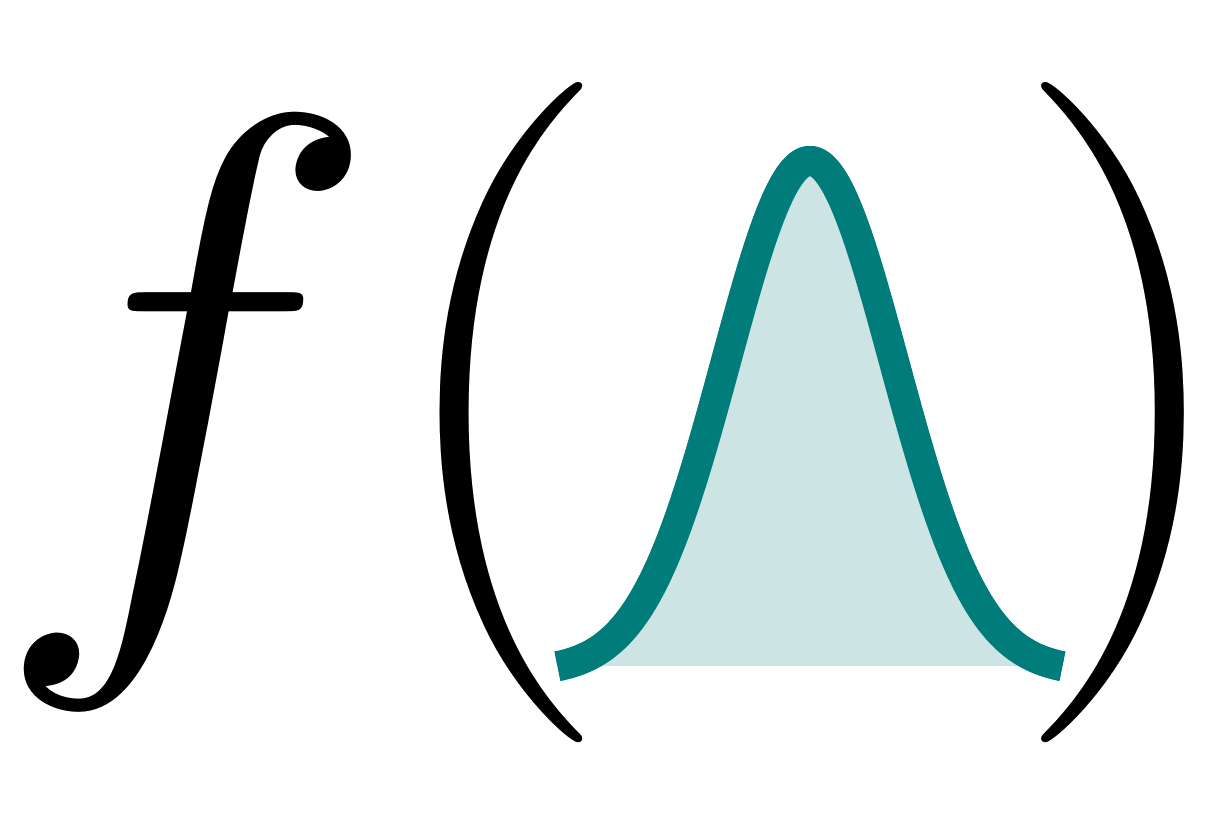}
	\end{split}
	\begin{split}\text{
			\url{https://probnum.readthedocs.io/}}.~~~~~~~~~~~~~~~~~~~~~~~~~~~
	\end{split}
\end{align*}
Future algorithmic improvements will likely end up there, too.

In summary:
the presented tricks effectively remove a barrier in computing probabilistic ODE solutions when it comes to high-order algorithms and small steps.
This allows using probabilistic ODE solvers as a drop-in replacement for other, high-order, rapidly converging, classical algorithms, thereby enriching chains of statistical computation that involve numerical simulation of dynamical systems with cheap yet effective uncertainty quantification---all of which is now possible without losing out on speed or reliability of the simulation.

\acks{
	The authors gratefully acknowledge financial support by the German Federal Ministry of Education and Research (BMBF) through Project ADIMEM (FKZ 01IS18052B).
	They also gratefully acknowledge financial support by the European Research Council through ERC StG Action 757275 / PANAMA; the DFG Cluster of Excellence “Machine Learning - New Perspectives for Science”, EXC 2064/1, project number 390727645; the German Federal Ministry of Education and Research (BMBF) through the T\"ubingen AI Center (FKZ: 01IS18039A); and funds from the Ministry of Science, Research and Arts of the State of Baden-W\"urttemberg.
	The authors thank the International Max Planck Research School for Intelligent Systems (IMPRS-IS) for supporting N. Kr\"amer.

	They are grateful to Nathanael Bosch and Filip Tronarp for many valuable discussions. {N. Bosch maintains a Julia implementation of probabilistic ODE solvers.\footnote{ \url{https://github.com/nathanaelbosch/ODEFilters.jl}.}}
	They further thank Hans Kersting, Jonathan Schmidt, Marius Hobbhahn, and Elizabeth Baker for helpful feedback on the manuscript.
}

\appendix
\section{Implementation Guide}
\label{appendix:implementation_guide}

The following explains detailed iteration schemes of the probabilistic ODE solver, including initialisation (\ref{appendix:initialisation}), prediction (\ref{appendix:prediction}), update (\ref{appendix:update}), and smoothing (\ref{appendix:smoothing}).

\subsection{Initialisation}
\label{appendix:initialisation}
Choose an order $\nu$ (recommendations were made in the discussion in Section \ref{sec:discussion}).
Initialise the ODE solver with Taylor-mode automatic differentiation. The covariance has zeros, respectively.
Before the first step, assemble $\overline{\vec{A}}$ and decompose $\overline{\vec{Q}}$ into its Cholesky factors, $\overline{\vec{Q}} = \vec{L}_Q \vec{L}_Q^\top$.
For high orders, this remains a numerical bottleneck, because even $\overline{\vec{Q}}$ is ill-conditioned for large $\nu$; recall Table \ref{table:taylor_nordsieck}.
If not all derivatives are initialised accurately, set the respective entries of the initial covariance $\vec{C}_0$ to a non-zero value and decompose it into its Cholesky factors, $\vec{C}_0 = \vec{L}_0 \vec{L}_0^\top$ (using the LDL decomposition if necessary).

\subsection{Prediction}
\label{appendix:prediction}
Mean and covariance are stored in the original, non-transformed coordinates. Therefore, the update step consists of (i) applying the transformation
\begin{align}
	\overline{\vec{m}}_n = \vec{T}^{-1}_{n} \vec{m}_n, \quad
	\overline{\vec{L}}_n = \vec{T}^{-1}_{n} \vec{L}_n
\end{align}
and (ii) computing the prediction in the changed coordinate system,
\begin{align}
	\vec{m}_{n+1}^-                                                         & = \overline{\vec{A}}\, \overline{\vec{m}}_n,                   \\
	\left(\overline{\vec{A}}\, \overline{\vec{L}}_n, \vec{L}_Q \right)^\top & = \vec{X} \,\vec{R}_{n+1}                 \label{eq:qr_decomp} \\
	\vec{L}_{n+1}^-                                                         & = (\vec{R}_{n+1})^\top_{0:d(\nu + 1)}
\end{align}
where Equation \eqref{eq:qr_decomp} is a QR decomposition.
$\vec{X}$ is discarded. The notation $(\vec{R}_{n+1})_{0:d(\nu + 1)}^\top$ implies that the top $d(\nu+1)\times d(\nu+1)$ block of $\vec{R}_{n+1}$ is extracted and transposed.

\subsection{Update}
\label{appendix:update}
The predicted mean and covariance ``live in the preconditioned space''.
The update consists of a measurement step and a conditioning step.
The measurement step starts with assembling either $\vec{H} = \vec{E}_1^\top \vec{T}_n$ (EK0) or $\vec{H} = \vec{E}_1^\top\vec{T}_n - \nabla f(\vec{E}_0^\top \vec{T}_n \vec{m}_{n+1}^-) \vec{E}_0^\top \vec{T}_n$ (EK1), and continues with computing
\begin{subequations}
	\begin{align}
		\vec{z}_{n + 1}               & = \vec{E}_1^\top \vec{T}_{n}\vec{m}_{n+1}^- - f(\vec{E}_0^\top\vec{T}_{n}\vec{m}_{n+1}^-) \\
		(\vec{H}\vec{L}_{n+1}^-)^\top & = \vec{X} \, \vec{R}_{n+1}^S                                                              \\
		\vec{L}_{S}                   & = (\vec{R}_{n+1}^S)^\top_{0:d(\nu + 1)}                                                   \\
		\vec{C}_{\text{cross}}        & = \vec{L}_{n+1}^- (\vec{L}_{n+1}^-)^\top \vec{H}^\top .
	\end{align}
\end{subequations}
Other than in Eq. \eqref{eq:transformed_equivalent_discretisation}, $\vec{T}_n$ is part of $\vec{H}$ now.
The conditioning step is
\begin{subequations}
	\begin{align}
		\vec{K}_{n+1}            & = \vec{C}_{\text{cross}} \vec{L}_{S}^{-\top} \vec{L}_{S}^{-1} \\
		\overline{\vec{m}}_{n+1} & = \vec{m}_{n+1}^- - \vec{K}_{n+1} \vec{z}_{n+1}               \\
		\overline{\vec{L}}_{n+1} & = ( \vec{I} -
		\vec{K}_{n+1} \vec{H})\vec{L}_{n+1}^- .
	\end{align}
\end{subequations}
Since each of $\vec{H}$, $\vec{T}_n$ and $\vec{L}_{n+1}^-$ are of full rank, $\vec{L}_S$ is invertible.
Inversion of $\vec{L}_S \vec{L}_S^\top$ leverages the readily computed Cholesky-decomposition.
After the respective update,
$\overline{\vec{m}}_{n+1}$ and
$\overline{\vec{L}}_{n+1}$ still ``live in the preconditioned space''. Therefore, they need to be transformed back to the original coordinates
\begin{align}
	\vec{m}_{n+1} = \vec{T}_{n} \overline{\vec{m}}_{n+1}, \quad
	\vec{L}_{n+1} = \vec{T}_{n} \overline{\vec{L}}_{n+1},
\end{align}
before storing them.
$\vec{L}_{n+1}$ is not necessarily triangular or positive definite, but since it is a matrix square-root of $\vec{C}_{n+1}$, the posterior covariance is guaranteed to be symmetric and positive semidefinite.

\subsection{Smoothing}
\label{appendix:smoothing}

First, all states are fetched into the ``preconditioned coordinate system'',
\begin{subequations}
	\begin{align}
		\overline{\vec{m}}_n^F     & = \vec{T}_{n}^{-1} \vec{\vec{m}}_n^F, \quad~~~~~
		\overline{\vec{L}}_n^F = \vec{T}_{n}^{-1} \vec{\vec{L}}_n^F,                  \\
		\overline{\vec{m}}_{n+1}^S & = \vec{T}_{n}^{-1} \vec{\vec{m}}_{n+1}^S, \quad
		\overline{\vec{L}}_{n+1}^S = \vec{T}_{n}^{-1} \vec{\vec{L}}_{n+1}^S,
	\end{align}
\end{subequations}
after which the prediction step is repeated (it has to be repeated only on paper, implementations can reuse predictions from the filtering recursion),
\begin{align}
	\vec{m}_{n+1}^-                                                           & = \overline{\vec{A}}\, \overline{\vec{m}}_n^F                       \\
	\left(\overline{\vec{A}}\, \overline{\vec{L}}_n^F, \vec{L}_Q \right)^\top & = \vec{X} \,\vec{R}_{n+1}                      \label{eq:second_qr} \\
	\vec{L}_{n+1}^-                                                           & = (\vec{R}_{n+1})^\top_{0:d(\nu + 1)},
\end{align}
and again, Equation \eqref{eq:second_qr} is a QR decomposition.
$\overline{\vec{A}}$ has full rank, $\overline{\vec{L}}_n^F$ is positive semidefinite, and $\vec{L}_Q$ is positive definite, therefore $\vec{L}_{n+1}^-$ is invertible.
Second, the update is computed as
\begin{subequations}
	\begin{align}
		\vec{G}_{n +1}                                                                                                              & = \overline{\vec{L}}_n^F \left(\overline{\vec{A}}  \, \overline{\vec{L}}_n^F \right)^\top \left(\vec{L}_{n+1}^-\right)^{-\top} \left(\vec{L}_{n+1}^-\right)^{-1} \\
		\overline{\vec{m}}_n^S                                                                                                      & = \overline{\vec{m}}_n^F - \vec{G}_{n+1} \left[\overline{\vec{m}}_{n+1}^S - \vec{m}_{n+1}^- \right]                                                              \\
		\left((\vec{I} - \vec{G} \overline{\vec{A}})\, \overline{\vec{L}}_n, \vec{G}\vec{L}_Q , \vec{G}\vec{L}_{n+1}^S \right)^\top & = \vec{X} \,\vec{R}_{n+1}                                                                                                                                        \\
		\overline{\vec{L}}_n^S                                                                                                      & = (\vec{R}_{n+1})^\top_{0:d(\nu + 1)}
	\end{align}
\end{subequations}
where the penultimate line is a QR decomposition that computes a Joseph-style update for the smoothing iteration; this is a counterpart to Eq. (4.23) in the book by \citet{grewal2014kalman}, applied to the smoothing step.
Finally, before storing the values, the results are pushed back to the original coordinate system,
\begin{align}
	\vec{m}_{n}^S = \vec{T}_{n} \overline{\vec{m}}_{n}^S, \quad
	\vec{L}_{n}^S = \vec{T}_{n} \overline{\vec{L}}_{n}^S.
\end{align}
This concludes the smoothing step.

We emphasise that at least on paper, the outcome of these steps is identical to the outcome of ODE filters and smoothers in the standard implementation. In practice, the results may differ, though, because of accumulation of round-off errors in the ``classical'' implementation.

\vskip 0.2in
\bibliography{bibfile.bib}

\end{document}